\pdfoutput=1

\documentclass[11pt]{article}
\usepackage{tabularx}
\usepackage{booktabs}
\usepackage{xcolor}
\usepackage{arydshln} 
\usepackage{multirow}
\usepackage{graphicx}
\usepackage{CJKutf8}
\usepackage{tcolorbox}
\usepackage{mdframed}
\usepackage{makecell}
\usepackage[final]{acl}

\usepackage{times}
\usepackage{latexsym}
\usepackage[T1]{fontenc}

\usepackage[utf8]{inputenc}

\usepackage{microtype}

\usepackage{inconsolata}

\newcommand{\ie}{\textit{i.e.}}
\newcommand{\eg}{\textit{e.g.}}

\newcommand{\methodname}{CLongEval}
\newcommand{\lllm}{long-context LLMs}

\usepackage{colortbl}
\usepackage{graphicx}
\usepackage{booktabs}
\usepackage{enumitem}
\usepackage{amssymb}
\usepackage{bbding}
\usepackage[utf8]{inputenc}
\usepackage{cleveref}
\crefname{section}{§}{§§}
\Crefname{section}{§}{§§}

\title{CLongEval: A Chinese Benchmark for Evaluating Long-Context Large Language Models}

\author{Zexuan Qiu$^{1}$\thanks{Equal Contribution.}, Jingjing Li$^{1*}$, Shijue Huang$^{2}$, Xiaoqi Jiao$^{3}$, Wanjun Zhong$^{4}$, Irwin King$^{1}$\thanks{Corresponding author.}\\
$^1$ The Chinese University of Hong Kong \quad $^2$ Harbin Institute of Technology (Shenzhen) \\
$^3$ LightSpeed Studios, Tencent Inc. \quad $^4$ Sun Yat-Sen University \\ 
{\{zxqiu22, lijj, king\}@cse.cuhk.edu.hk }
 }
\begin{document}
\maketitle
\begin{abstract}
Developing Large Language Models (LLMs) with robust long-context capabilities has been the recent research focus, resulting in the emergence of \lllm~ proficient in Chinese.
However, the evaluation of these models remains underdeveloped due to a lack of benchmarks.
To address this gap, we present \methodname, a comprehensive Chinese benchmark for evaluating \lllm. \methodname~is characterized by three key features: (1) Sufficient data volume, comprising 7 distinct tasks and 7,267 examples; (2) Broad applicability, accommodating to models with context windows size from 1K to 100K; (3) High quality, with over 2,000 manually annotated question-answer pairs in addition to the automatically constructed labels.
With \methodname, we undertake a comprehensive assessment of 6 open-source \lllm~and 2 leading commercial counterparts that feature both long-context abilities and proficiency in Chinese.
We also provide in-depth analysis based on the empirical results, trying to shed light on the critical capabilities that present challenges in long-context settings.\footnote{The dataset, evaluation scripts, and model outputs are released in \url{https://github.com/zexuanqiu/CLongEval}.}

\end{abstract}

\vspace{-1mm}

\section{Introduction}
Large Language Models have demonstrated impressive performance across a wide range of Natural Language Processing~(NLP) tasks, including machine translation~\cite{hendy2023good, jiao2023chatgpt}, fact checking~\cite{DBLP:journals/corr/abs-2305-14225},  text style transfer~\cite{reif-etal-2022-recipe, li2020unsupervised} and other generation tasks~\cite{DBLP:journals/corr/abs-2310-20352,li2022text}. 
To enable LLMs to support more intricate and diverse applications, an increasing number of studies focus on extending the context window these models can handle. Consequently, many \lllm~that support Chinese have emerged, including both commercial models~\cite{GPT4turbo} and open-source ones~\cite{zhllama, bai2023qwen, chatglm3, internlm2}, of which the context lengths span from 32K to 200K. Despite these developments, the efficacy of models in long-context settings remains underexamined, primarily due to the lack of a robust evaluation benchmark.

\begin{figure}
    \centering
    \includegraphics[width=0.9\linewidth]{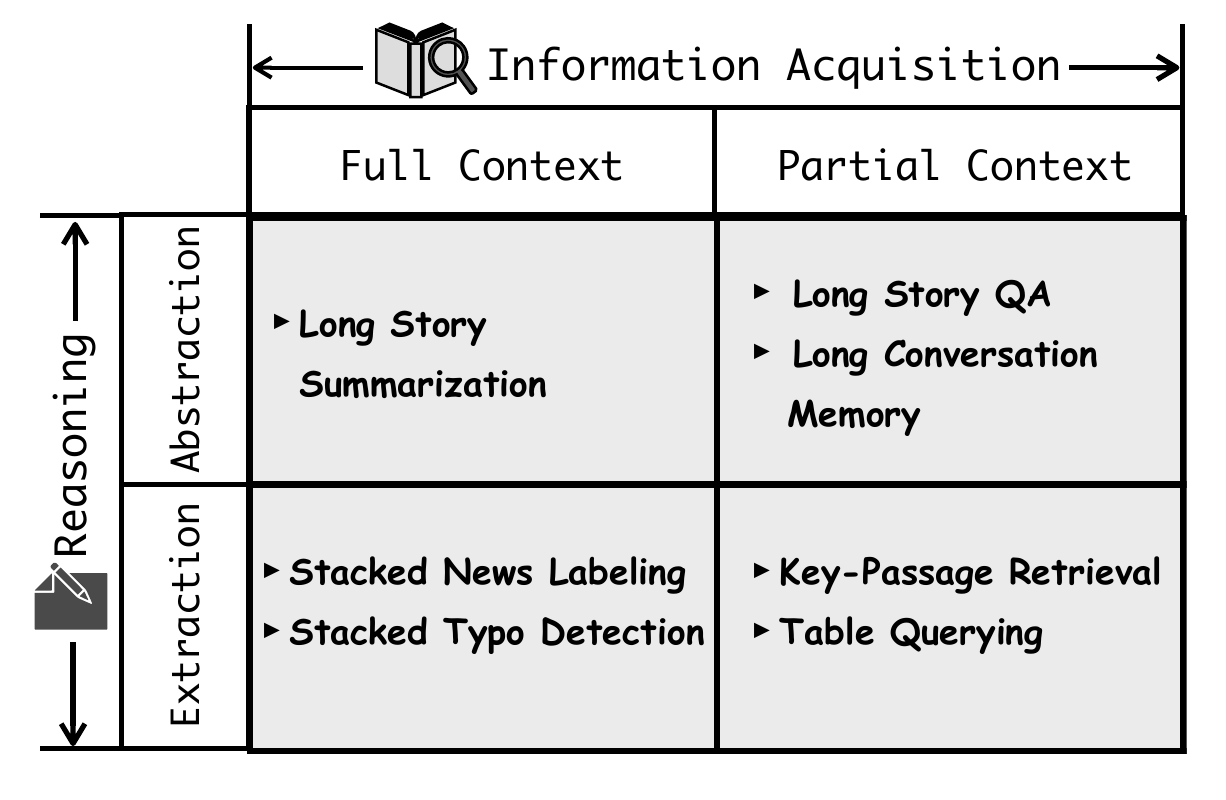}
    \caption{The evaluation framework of \methodname. The seven test tasks in \methodname~are designed to comprehensively assess two important capabilities of \lllm: information acquisition and reasoning. }
    \label{fig:framework}
\end{figure}

Recently, a few benchmarks have been proposed for the evaluation of English \lllm~\cite{an2023leval, bai2023longbench}. As for the Chinese domain, there exists only a bilingual benchmark~\cite{bai2023longbench}, wherein only 5 out of 21 test tasks are designed for Chinese. This benchmark offers only 1K instances in total, with an average token length capped at approximately 13K\footnote{The reported average length is 13,386 characters, yielding at most 13K tokens after tokenization.}, rendering it inadequate for a comprehensive evaluation. Therefore, there is an urgent need for a high-quality Chinese benchmark for \lllm. Considering the considerable advancements in this field, the establishment of such a benchmark facilitates a thorough investigation of existing models, which might bring insights to the research community.


In this paper, we present \methodname, a benchmark designed for evaluating Chinese \lllm. Prior to the construction of datasets, we conduct a systematic analysis of the key capabilities requisite for handling long context,
to ensure a thorough assessment of the model's functionality~(\cref{sec:framework}).  Analogous to the human problem-solving paradigm, the basic functionalities of \lllm~can be conceptualized as: (1) the capacity to precisely identify and acquire the key information framing in either \textit{partial} or \textit{full} context; and (2) 
the competence to reason out the answer based on the given information in either an \textit{extractive} or \textit{abstractive} manner.
These key abilities establish the evaluation framework behind \methodname, which is illustrated in Figure~\ref{fig:framework}.



To accommodate models with varying spans of context windows, we consider three subsets within \methodname: small set (1K-16K), medium set (16K-50K), and large set (50K-100K)~(\cref{sec:data_config}). 
In dataset construction, we select test tasks that correspond to the capabilities outlined in the evaluation framework. Moreover, we ensure that primary test tasks are highly aligned with real-life user scenarios so that the benchmark can accurately reflect models' capability in practical applications~\cite{longllama2}~(\cref{sec:data_construct}). 
Overall, we craft 7 distinct tasks in \methodname: 2 tasks are human-annotated, 1 task is GPT-4-annotated and 4 tasks are re-constructed from public datasets.


With \methodname, we evaluate 8 \lllm~proficient in Chinese, including two commercial models known for their powerful long-text processing capability: Moonshot-v1-128K and GPT-4-Turbo-128K~(\cref{sec:exp_main}). We highlight several key findings in the long-context setting: (1) The commercial models consistently outperform open-source models across tasks, and the performance gap is particularly evident in tasks that primarily involve straightforward information extraction. (2) Extraction with full context is the most challenging setting. GPT-4-Turbo displays a more significant decline in performance as context length increases, compared to other settings. (3) For tasks requiring an understanding of partial context, the answer's position within a long context does not consistently lead to significant performance fluctuations. More analysis and observations are elaborated in~\cref{sec:exp_analysis} and Appendix~\ref{app:analysis}.

\vspace{-1mm} 
\section{Evaluation Framework in \methodname}\label{sec:framework}
To offer a thorough and systematic evaluation, we analyze the key capabilities necessary for the efficacy of \lllm.
Generally speaking, the capacity for a long-context LLM in interpreting human textual instructions largely depends on its ability of \textit{information acquisition}. Moreover, an indispensable ability for these models extends beyond mere information collection to encompass \textit{reasoning} based on the assimilated information.

\paragraph{Long-Context Information Acquisition.} It refers to the capability to recognize and parse relevant information framed in extensive and complex textual input. It bottlenecks LLMs' effectiveness in synthesizing its contextualized knowledge to execute a wide array of tasks, from answering questions to carrying out complex instructions. Moreover, as the length of the input text increases, maintaining a coherent and precise grasp of the input information becomes increasingly challenging.

In this evaluation dimension, we conduct a two-fold classification based on the distribution of information requisite for task fulfillment: full-context and partial-context information acquisition. For each category, we introduce an array of test tasks tailored to assess the model's proficiency, including tasks that demand an accurate comprehension of the entire input (\ie~full-context), and those that rely on a correct understanding of selective snippets of the input (\ie~partial-context), respectively. Figure~\ref{fig:framework} illustrates the categorization of tasks that require either partial- or full-context in information acquisition.

\paragraph{Long-Context Reasoning.} It refers to the ability to perform the inferential process of synthesizing a conclusion from presented lengthy statements~\cite{pi2022logigan}.
In real-world applications, most tasks require not only a precise understanding of the input text but also the capacity of reasoning based on the provided information. LLMs equipped with proficient reasoning abilities can navigate the problem-solving and decision-making procedure, both of which are crucial cognitive functions necessary for handling complex tasks~\cite{pomerol1997artificial}.

In the reasoning process, outputs can be synthesized through two ways: content extraction and abstraction. Accordingly, the evaluation dimension of reasoning incorporates these two distinct types of test tasks.
The abstraction tasks involve generating new content not explicitly in the source material, demanding a deeper understanding and recombination of input. In contrast, extraction tasks assess the model's ability to directly identify and extract information from the input without altering the original content. This framework enables a nuanced evaluation of LLMs' reasoning capabilities, including the capacity for generating novel insights and accurately retrieving information.
Figure~\ref{fig:framework} illustrates the tasks to test the reasoning ability in either an extractive or abstractive manner. 

%



\begin{table*}[!t]
\centering
\setlength{\tabcolsep}{1mm}{
{\fontsize{9pt}{14pt} \selectfont
\begin{tabularx}{\linewidth}{l c c >{\centering\arraybackslash}X>{\centering\arraybackslash}X>{\centering\arraybackslash}X >{\centering\arraybackslash}X>{\centering\arraybackslash}X>{\centering\arraybackslash}X >{\centering\arraybackslash}X>{\centering\arraybackslash}X>{\centering\arraybackslash}X}
\toprule
\multirow{2}{*}{\textbf{Task Name}} & \multirow{2}{*}{\textbf{Annotated}} &\multirow{2}{*}{\textbf{\#Data}} & \multicolumn{3}{c}{\textbf{Small Set}} & \multicolumn{3}{c}{\textbf{Medium Set}} & \multicolumn{3}{c}{\textbf{Large Set}}  \\

 &  &  &Min &Max &\#Data &Min &Max &\#Data &Min &Max &\#Data \\
\midrule

Long Story QA     &\checkmark &995 
&1,693  &16,224  &294  &14,437  &50,553  &398  &49,636  &99,038  &299    \\ 
Long Conversation Memory &\checkmark     &1,067  
&768 &15,589 &358 &14,669 &43,225 &353 &41,481 &88,731 &356  \\  
Long Story Summarization &$\times$     &1,000  
&1,022 &13,716 &300 &12,958 &48,677 &400 &42,046 &87,092 &300   \\ 
Stacked News Labeling  &$\times$ &1,005  
&738 &13,521 &303 &13,635 &44,100 &402 &43,875 &84,475 &300     \\ 
Stacked Typo Detection  &$\times$     &1,000     
&1,069 &16,385 &550 &15,956 &51,016 &300 &51,404 &98,803 &150  \\ 
Key-Passage Retrieval   &$\times$     &1,100      
&1,249 &17,830 &400 &18,006 &56,086 &400 &55,367 &95,073 &300   \\ 
Table Querying    &$\times$     &1,100  
&1,773 &24,597 &400 &23,894 &74,004 &400 &73,984 &125,529 &300  \\ 
\bottomrule
\end{tabularx}
\vspace{-2mm}
\caption{An overview of the test tasks in \methodname. \textbf{Annotated} denotes whether the test samples are newly human-annotated. Min and Max refer to the minimum and maximum lengths of the examples within each subset.}
\label{tab:clongeval}}
}
\end{table*}
\vspace{-2mm}



\section{The \methodname~Benchmark}
\vspace{-1mm}

\subsection{Dataset Configuration}\label{sec:data_config}
Anchored by the capabilities outlined in the evaluation framework, we create \methodname,  which contains 7,267 test samples across 7 tasks, including  Long Story QA, Long Conversation Memory, Long Story Summarization, Stacked News Labeling, Stacked Typo Detection, Key-Passage Retrieval and Table Querying. Among them, Long Story QA and Long Conversation Memory are human-annotated, Long Story Summarization is GPT-4-annotated and the rest 4 tasks are re-constructed from public datasets.
An overview of all tasks in \methodname~and detailed statistics are provided in Table~\ref{tab:clongeval}. In this paper, we use  InternLM2~\cite{internlm2} tokenizer to tokenize the input and report the number of tokens as context length.

We notice the divergence of context lengths supported by existing \lllm. To ensure a broad scope of applicability of \methodname, we stratify the benchmark into three subsets: a small set, a medium set, and a large set. 
Specifically, the small set primarily includes test data with lengths ranging from 1K to 16K tokens, the medium set mainly encompasses lengths from 16K to 50K tokens, and the large set primarily extends from 50K to 100K tokens.



\vspace{-2mm}

\subsection{Dataset Construction}\label{sec:data_construct}


For a comprehensive evaluation, 7 tasks are collected in alignment with the predefined evaluation framework. The examples for each task are provided in Appendix~\ref{sec:data_sample}.
\paragraph{Long Story QA (LStQA)}
The Long Story QA task involves LLMs answering questions based on a context snippet from a long story. To excel at this task, the model should identify the relevant snippet and abstractively reason out the answer. Unlike the normative and objective nature of MultiFieldQA~\citep{bai2023longbench}, the stories we choose are narrative, creative, and inherently longer, offering a valuable addition to lengthy single-document QA evaluation. Inspired by NarrativeQA~\citep{kovcisky2018narrativeqa}, this task involves annotated questions that refer to non-consecutive parts of the text. Annotators are encouraged to provide concise answers in their own words rather than copying directly from the snippet.

We curate 153 Chinese narrative novels from a  website\footnote{\url{https://www.wenshuoge.com}.} that gathers public domain books. The collected novels cover genres including martial arts, social themes, and mysteries. 200 non-overlapping stories are extracted from the collection, and the number of questions per story is proportional to its length in tokens, resulting in more questions for longer stories.  We then extract snippets from each story, evenly distributed throughout, with an average of 720 Chinese characters per snippet. The number of snippets for each story corresponds to the expected number of questions. Then 6 question-answer pairs for each snippet are generated by instructing GPT-4-Turbo following the aforementioned annotating principles.  Annotators then select a specific question-answer pair that is most related to a given snippet from 6 options. In addition, it is ensured that the complexity of questions related to characters, events, and the reasons behind occurrences is maintained. The questions are manually revised to include chapter information and replace pronouns with character names, making the questions more specific. There are 995 question-answer pairs based on 200 stories (\ie, contexts), with an average question length of 18.5 Chinese characters and an average answer length of 11.0 Chinese characters.  Note that snippets are used for annotation, whereas during testing, the model is still required to find answers from the entire story.

\paragraph{Long Conversation Memory (LCvMem)}
This task is designed to assess a model's long-term memory capability. This task utilizes inputs from multi-day conversations between a user and a companion chatbot, where the model is required to accurately respond to questions about specific details from the conversation history of a particular day. 
It determines the ability to maintain contextual understanding, ensure meaningful interactions, and interpret user behaviors over time~\cite{zhong2023memorybank}. Moreover, this capability becomes increasingly crucial as the length of the model's input extends, presenting a greater challenge in retaining a precise memory of the input content.


To construct the test dataset, we utilize dialogue records from 140 days of interactions between 80 virtual users and companion chatbots, and manually annotate 1,067 QA pairs. We adopt the experimental setting in~\citet{zhong2023memorybank} to construct the evaluation dataset. 
For the user profiles, we manually craft profiles for 20 virtual users, including names, personalities, and topics of interest, and prompt GPT-4-Turbo to generate the rest virtual user profiles. 
Leveraging the user meta-information, we employ GPT-4-Turbo to simulate dialogues between different users and companion chatbots in 140 days. Due to the limitation of context windows, we apply the hierarchical event summary in~\citet{zhong2023memorybank} to generate long dialogues.
All the generated conversation records are reviewed and deduplicated. 
Given the dialogue records, we manually craft 1,067 probing questions and answers to evaluate the model's ability to accurately retrieve relevant memories and generate appropriate responses. 


\paragraph{Long Story Summarization (LStSum)}

Text summarization is to distill information from a source text and present it in a condensed form. 
As a pivotal task in natural language processing, summarization requires a full-context understanding of input and complex reasoning.
In \methodname, we introduce a long story summarization task that comprises long-context input based on the story, which is more practical needs and poses more challenges in the aggregation of long-context information. 

To obtain high-qulity long-context Chinese corpus, we utilize Cloud Translation API\footnote{\url{https://cloud.google.com/translate}.} to translate the BOOKSUM dataset~\citep{kryściński2022booksum} into Chinese, which covers books from various domains and includes highly abstractive, human written summaries on three levels: paragraph, chapter-, and book-level. Formally, each sample $(t_i, s_i)$ in BOOKSUM comprises a textual input $t_i$ and its corresponding summary $s_i$, and $t_i$ may be a paragraph, a chapter, or a whole book.
We choose continuous paragraphs or chapters $[t_i, t_{i+1}, ..., t_j]$ in expected length and concatenate them to construct long-context input $T$ to ensure coherent semantics from the translated BOOKSUM dataset. Subsequently, we utilize GPT-4-Turbo to aggregate the corresponding summaries $[s_i, s_{i+1}, ..., s_j]$ of the chosen continuous paragraphs or chapters into an overall summary $S$, which can be regarded as the appropriate and highly abstractive summary of the constructed long-context input.
All the generated summary $S$ is passed to manual check and refinement to guarantee the quality.

\paragraph{Stacked News Labeling (StNLab)}
In this task, $N$ news articles are stacked in one single context, with each article containing a news index (ranging from 1 to N) and its content. The goal of this task is to assess whether the LLMs can comprehensively read all news articles in a long context and determine the category of each news from given possible category pools at once.  Completing this task requires the model to carefully read and analyze all the information within the long context. This task is akin to \textit{SpaceDigest} of ZeroScrolls~\citep{shaham2023zeroscrolls} or \textit{PassageCount} of LongBench~\citep{bai2023longbench}, where LLMs analyze long contexts piece by piece. However, their requested outputs are aggregated numbers (\eg, the count of positive reviews), making it difficult to gauge the LLMs' genuine understanding of each part. The proposed stacked news labeling task, in contrast, presents a more demanding challenge that tests the ability to comprehend lengthy contexts.

To construct this dataset, we begin by extracting a subset from THUnews~\citep{thunews2016}, an extensive collection of around 840K Chinese news articles.The subset include 9 categories: \textit{Sports}, \textit{Entertainment}, \textit{Home}, \textit{Real Estate}, \textit{Education}, \textit{Politics}, \textit{Gaming}, \textit{Technology}, and \textit{Finance}.  Each category contains an equal number of news articles, with the sampled articles having an average Chinese character count of 588.1. We randomly select news articles from different categories to fill the context until the desired context length was reached.  Finally, we create 1005 contexts as test samples.

\paragraph{Stacked Typo Detection (StTDet)} 
Typo Detection is aimed at extracting misspelled Chinese characters from a given input. Unlike prior works~\citep{tseng2015introduction,lv2023general} that focus on sentence-level typo recognition, our tacked typo detection aims to identify all typos present in the lengthy input, which is of practical importance. This task requires LLMs to have full-context understanding capabilities as well as distinguished information extraction abilities.

cWe utilize the collected Chinese narrative corpus same as Long Story QA to generate 1000 contexts as test samples. Each context is divided into multiple paragraphs, identified by a paragraph ID that starts from 0 and increases incrementally. We randomly select some paragraphs and choose one Chinese character as a typo candidate from each selected paragraph. A corresponding homophone is then used to replace the chosen character, creating a homophonic typo. To maintain a balanced distribution of typos, the number of typos is determined based on the data length: 10 for the small set, 20 for the medium set, and 30 for the large set. Roughly half of the paragraphs in each context contain misspelled characters.

\begin{table*}[!ht]
    \centering
    {\fontsize{10pt}{13pt} \selectfont
    \begin{tabularx}{\textwidth}{lXXXXXXX}
    \specialrule{1.3pt}{0pt}{0pt}
    \textbf{Model}   & LStQA  & LCvMem   & LStSum    &StNLab  &StTDet   &KpRet & TblQry \\
    \midrule
    \multicolumn{8}{c}{\textbf{Small Set}} \\
    Zh-LLAMA2-7B-64K   &29.34  &41.10  &10.29  &0.59  &0   &2.86   &7.50 \\
    Zh-Alpaca2-7B-64K  &35.52  &29.34  &14.29  &4.97  &0.09    &6.39   &9.75 \\ 
    Qwen-7B-32K             &31.94  &47.71  &11.20  &4.31   &0      &11.18  &6.64 \\ 
    ChatGLM3-6B-32K         &49.36  &53.40  &16.37  &0.46   &0.91   &33.67  &22.60 \\ 
    InternLM2-7B-32K        &49.55  &58.34  &17.29  &16.46  &2.27   &21.87  &20.75 \\ 
    InternLM2-20B-32K       &53.82  &57.41  &17.00  &11.16  &0.91   &34.97  &17.25 \\
    Moonshot-v1-32K            &60.21      &51.76      &21.56      &\textbf{89.01}      &25.36      &\textbf{86.74}      &66.50     \\
    GPT-4-Turbo-128K         &\textbf{66.19}  &\textbf{63.42}  &\textbf{21.96}  &79.70  &\textbf{38.35}  &84.24  &\textbf{82.35} \\ 
    \midrule
    \multicolumn{8}{c}{\textbf{Medium Set}} \\
    Zh-LLAMA2-7B-64K   &16.90  &26.30  &7.74   &0  &0      &1.21   &N/A    \\ 
    Zh-Alpaca2-7B-64K  &18.41  &22.45  &8.56   &0  &0      &0.93   &N/A    \\
    InternLM2-7B-200K   &29.59  &32.07  &8.13   &0  &0    &1.45   &4.50   \\
    InternLM2-20B-200K  &25.13  &36.84  &13.99  &0  &0    &1.64   &6.25   \\
    Moonshot-v1-128K           &51.20     &38.29      &\textbf{18.81}      &\textbf{86.30}      &\textbf{11.33}      &\textbf{78.64}      &\textbf{66.50 }    \\
    GPT-4-Turbo-128K         &\textbf{52.63}  &\textbf{54.18}      &17.38    &37.40  &9.32   &22.40      &52.76      \\
    \midrule
    \multicolumn{8}{c}{\textbf{Large Set}} \\
    InternLM2-7B-200K  &19.03  &18.16  &2.36   &0  &0    &0.89   &2.67   \\ 
    InternLM2-20B-200K &15.62  &28.39  &8.31   &0  &0    &0.51   &0.67   \\
    Moonshot-v1-128K           &\textbf{41.52}  &\textbf{32.59}  &\textbf{16.38}   &\textbf{78.48}    &\textbf{4.33}  &\textbf{51.50}                 &\textbf{52.00}   \\
    \specialrule{1.3pt}{0pt}{0pt}
    \end{tabularx}
    }
    \vspace{-2mm}
    \caption{The \methodname~Leaderboard. The results are up to date as of 02/15/2024. N/A means the maximum token length of the dataset surpasses the model's context window. Zh-LLAMA2/Alpaca2 denotes Chinese-LLAMA2/Alpaca2 for short.}
    \vspace{-3mm}
    \label{tab:main}
\end{table*}

\paragraph{Key-Passage Retrieval (KpRet)}

In this synthetic key-passage retrieval task, the context comprises a JSON object serialized as a string, containing multiple key-passage pairs. Each key is a unique string of 32 randomly generated characters including both letters and numbers, while the corresponding value is a continuous passage in Chinese. The objective of this task is to retrieve the corresponding passage directly based on the given key. 
Unlike \textit{LStQA} and \textit{LCvMem}, \textit{KpRet} focuses on the model's information extraction ability, rather than summarizing the answer from a located snippet. \textit{KpRet} draws inspiration from the synthetic key-value retrieval task mentioned in~\citep{liu2023lost}, but differs in that we aim to provide semantically meaningful natural language text instead of randomly generated 128-bit UUID strings, aligning more closely with real-world scenarios of passage retrieval~\citep{nguyen2016ms}. The main challenge lies in accurately retrieving and reproducing relatively long passages in their entirety.

All passages are sampled from three Chinese QA datasets, namely WebQA~\citep{li2016dataset}, Sogou QA\footnote{\url{https://github.com/sherlcok314159/ChineseMRC-Data}.}, and CMRC2018~\citep{cui2018span}, ensuring no repetition among them. All the passages exhibit a relatively consistent length, with an average of 81.2 Chinese characters. To construct 200 contexts, we have generated a substantial number of key-passage pairs. For each context, we uniformly select 5 questioned keys according to the position, resulting in a total of 1000 test examples.
\vspace{-1mm}
\paragraph{Table Querying (TblQry)}
In the table querying task, a context consists of multiple tables formatted in Markdown. In table querying, the objective is to locate a specific table within the context and retrieve a value from that table based on querying conditions.  Unlike the key-value data structure in \textit{KpRet}, \textit{TblQry} involves the model's simultaneous utilization of both row and column indices to extract a specific value from the table. Our question format follows a conditional pattern: "\textit{In Table A, when the value of Column B is C, what is the value of Column D?}” In this question format, LLMs need to first identify Table A among multiple tables in the context, then locate the row based on the value of Column B and retrieve the value of Column D. Moreover, Unlike \textit{KpRet} which returns long passages, \textit{TblQry} typically returns shorter values like numbers or names, with an average token length of 5.0. Therefore, this task primarily assesses LLMs' advanced contextual querying abilities rather than their proficiency in reproducing complex passages.


All the tables used in this task are sourced from WikTable~\citep{zhong2017seq2sql}, a collection of English tables. We filter out excessively long tables and retain only those with a token count not exceeding 2000 tokens.  Due to resource constraints, we only translate the column headers and the conditioned column into Chinese using the Cloud Translation API for each table. This ensures that the questions are posed in Chinese, while the returned values from the tables remain in English or numerical format. In total, 180 contexts containing multiple tables are constructed. The number of questions for each context is proportional to the number of tables it possesses, and we evenly distribute the tables that need to be queried across different positions within each context. Finally, 1100 test samples are generated.

\vspace{-2mm}
\section{Experiments}\label{sec:exp}

 \vspace{-1mm}

\subsection{Baselines}
8 LLMs are selected for evaluation based on whether they feature long context capability and exceptional support for the Chinese.
\textbf{Commercial Models}: (1) GPT-4-Turbo-128K, the \textit{GPT-4-1106-preview} model~\citep{GPT4turbo} with a 128K context window from OpenAI. (2) Moonshot-v1~\footnote{\url{https://platform.moonshot.cn}.} supporting up to 200K Chinese characters, developed by Moonshot AI. We call the 32K version to run the small set and call its 128K version for both medium and large sets. \textbf{Open-source Models}: (3) Chinese-LLAMA2-64K~\citep{zhllama}, extending context length of Chinese-LLAMA2 to 64K via YaRN~\cite{peng2023yarn} . (4) Chinese-Alpaca2-64K~\citep{zhllama}, the 64K context version of Chinese-Alpaca2. (5) Qwen-7B-32K~\citep{bai2023qwen}, extending Qwen-7B to 32K context length via NTK-aware scaled RoPE~\citep{ntkawarerope}. (6) ChatGLM3-6B-32K~\citep{chatglm3}, the 32K context version of ChatGLM3-6B. (7) InternlM2-7B-200K~\citep{internlm2}, effectively supporting ultra-long contexts of up to 200K tokens using dynamic NTK extrapolation~\citep{liu2023scaling}. (8) InternLM2-20B-200K~\citep{internlm2}, similar to InternLM2-7B-200K but is more robust and capable of handling intricate scenarios. For InternLM2-7B/20B with the small set, we rely on its native support for a 32K context window.  The values of the maximum output token limit for each task under different subsets are listed in Table~\ref{tab:max_tokens}. All the experiments are run on a server with 4 NVIDIA A100 (80GB) GPUs.


\vspace{-2mm}
\subsection{Evaluation Metrics} The evaluation is fully automatic. For \textit{LstQA} and \textit{LCvMem}, \textbf{F1} is employed to measure the unigram overlap between the generated and reference answer after ignoring white spaces and punctuation. For \textit{LstSum}, we use \textbf{ROUGE-L}~\citep{lin2004rouge} to measure the n-gram overlap between the generated and reference summary. For both \textit{StNlab} and \textit{StTDet}, a metric called \textbf{Average Accuracy} is introduced. It measures the ratio of the number of segments correctly answered by the generated answer to the total number of segments in the gold reference. On \textit{StNLab}, it indicates the percentage of news in the context that is correctly classified, while on \textit{StTDet}, it denotes the accuracy of identifying misspelled words in the context. For \textit{KpRet}, \textbf{Edit Score} based on Levenshtein distance is employed to measure the difference between the generated string and the gold reference string.  For \textit{TblQry}, \textbf{Exact Match} is utilized to measure whether the generated column value is identical to the gold reference.
For each of the 7 tasks, we first calculate the score per test sample using the aforementioned corresponding metrics and report the mean score across samples.


\vspace{-2mm} 
\subsection{Main Results}\label{sec:exp_main}
\vspace{-1mm}

\begin{table}[!t]
    \centering
    {\fontsize{9pt}{13pt} \selectfont
    \begin{tabularx}{\linewidth}{lXXX}
    \toprule
    \textbf{Model}   &\textbf{LStQA}  &\textbf{LCvMem}   &\textbf{LStSum}  \\
    \midrule
    Baichuan2-7B-4K     &23.18  &42.71  &7.88   \\
    Mistral-7B-8K       &23.12  &20.85  &8.95   \\
    Yi-6B-4K            &30.72  &27.29  &11.06  \\
    InternLM2-7B-4K     &35.52  &44.20   &16.44  \\
    InternLM2-7B-8K     &45.09  &52.64  &16.79  \\
    \midrule
    \midrule
    InternLM2-7B-32K    &49.55  &58.64   &17.29  \\
    \bottomrule
    \end{tabularx}}
    \vspace{-3mm}
    \caption{Peformance (\%) under truncated context in the small set.}
    \vspace{-5mm}
    \label{tab:truncate}
\end{table}

Table~\ref{tab:main} presents the performance on all datasets in CLongEval. We observe the following key findings from the experimental results: (i) For \textit{LstQA} and \textit{LCvMem}, GPT-4-Turbo does not show significant F1-score improvement compared to the top-performing open-source InternLM2-20B in the small set. However, it significantly outperforms InternLM2-20B (\eg, scoring 54.18 vs 36.84 on \textit{LCvMem}). Also, Moonshot-v1  exhibits less noticeable score degradation on medium and large sets compared to open-source models on \textit{LStQA}. (ii) For \textit{LStSum}, both Moonshot-v1 and GPT-4-Turbo show consistent Rouge-L scores on small and medium sets. (iii) For \textit{StNlab} and \textit{StTDet} which require careful analysis of full-text chunks to output either labeling results or identify spelling errors, there is a substantial performance gap between open-source and closed ones, with scores of all evaluated open-source models in the medium set being zero. GPT-4-Turbo's performance drops by 51.8\% when moving from the small set to the medium one on \textit{StNLab}. Meanwhile, Moonshot-v1 performs well on \textit{StNLab}, with only an 11.83\% decrease when expanding from the small set to the large one. (iv) For \textit{KpRet} and \textit{TlbQry} which involve information retrieval, all open-source models experience a sharp decline in performance as the input length increases, and Moonshot-v1 shows more robust handling of longer inputs compared to GPT-4-Turbo.

\begin{table}[!t]
    \centering
    {\fontsize{9pt}{14pt}\selectfont
    \begin{tabularx}{\linewidth}{cXXXX}
    \toprule
    \textbf{Model}  &\textbf{StNlab}    &\textbf{Nlab}  &\textbf{StTDet}    &\textbf{TDet} \\
    \midrule
     Qwen-7B         &4.31   &80.91  &0  &18.67 \\ 
     Zh-Alpaca2-7B   &4.97   &60.09  &0.09    &22.27 \\
     ChatGLM3-6B     &0.46   &86.71  &0.91    &34.23 \\
     InternLM2-7B    &16.46  &85.87  &2.27    &56.20 \\ 
     InternLM2-20B   &11.16  &84.04  &0.91    &56.90 \\
     Moonshot-v1     &\textbf{89.01}      &86.71           &25.36       &62.06      \\
     GPT-4-Turbo      &79.70  &\textbf{90.31}      &\textbf{38.22}   &\textbf{75.63}     \\ 
    \toprule
    \end{tabularx}}
    \vspace{-2mm}
    \caption{Perfomance (\%) comparison of \textit{StNLab} vs. \textit{NLab} and \textit{StTDet} vs. \textit{TDet}.}
    \vspace{-3mm}
    \label{tab:distraction}
\end{table}
Table~\ref{tab:truncate} presents the performance of smaller context window models Baichuan2-7B~\citep{yang2023baichuan}, Mistral-7B~\citep{jiang2023mistral} and Yi-6B~\footnote{\url{https://github.com/01-ai/Yi}.} on \textit{LstQA}, \textit{LCvMem}, and \textit{LstSum}. We also examine how the input length affects the performance of InternLM2-7B by truncating the context to 4K and 8K. Notably, shorter maximum context lengths result in lower scores, highlighting the need for effective long-context modeling in our benchmark.

\vspace{-2mm}
\subsection{Analysis}\label{sec:exp_analysis}
\begin{figure*}
    \centering
    \includegraphics[width=1.0\textwidth]{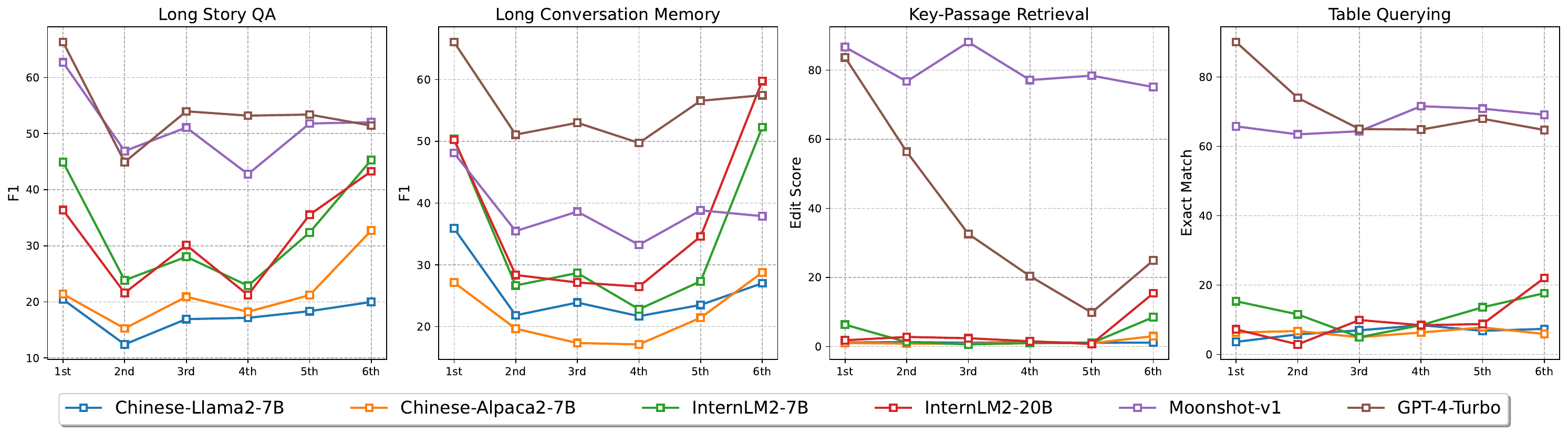}
    \caption{Effects of referenced chunk positions. The analyzed samples range from 10K to 60K in context length.}
    \vspace{-3mm}
    \label{fig:lost_in_middle}
\end{figure*}

\paragraph{Performance w.r.t. Answer Position}
We study how the position of the referenced chunk in the context affects the model's performance for four tasks that only need partial context. The results are shown in Figure~\ref{fig:lost_in_middle}. The position of the referenced chunk in the context is discretized into six intervals, with larger numbers indicating a closer position to the end. It is observed that for  \textit{LStQA} and \textit{LCvMem}, the evaluated models show a "lost in the middle" phenomenon~\citep{liu2023lost} where the models' performance decreases when the referenced chunks are in the middle of the context. For \textit{KpRet}, most open-source models only show some non-zero performance when the answer is located at the end of the context; GPT-4-Turbo shows a nearly linear decline in performance as the answer's position in the context becomes deeper, while Moonshot-v1 does not exhibit significant degradation. Similarly, for \textit{TblQry}, GPT-4-Turbo's performance drops as the answer's position goes deeper,  eventually getting surpassed by Moonshot-v1. The performance across different positions on \textit{KpRet} and \textit{TblQry} does not exhibit a distinct pattern.


\paragraph{Performance Discrepancy on \textit{StNlab} and \textit{StTDet}}
We are interested in evaluating LLMs' performance for the tasks of StNLab and StTDet, without considering the stacked long-context scenario. For the news labeling task (\textit{NLab} for short), we create a test set of 4,500 samples by sampling 500 news articles from each of 9 news categories on \textit{StNLab}, and each time a news article is given as input for the LLM to determine its category. For the typo detection task (\textit{TDet} for short), 3,000 paragraphs from \textit{StTDet} are sampled as test samples with each containing a typo, and LLMs are asked to take each paragraph as a model input to identify typos.
 Table~\ref{tab:distraction} reveals that open-source models achieve over 80\% accuracy for news labeling and at least 18\% accuracy for typo detection.  However, when news articles or paragraphs containing typos are stacked to form longer texts, their accuracy drops dramatically, even reaching 0 in the medium set as shown in Table~\ref{tab:main}.  Also, detailed performance results on \textit{StNLab}  in Appendix~\ref{sec: heat_map} illustrate that  GPT-4-Turbo misclassifies a significant portion of the news articles as the context length increases. In contrast, Moonshot-v1 consistently achieves high accuracy scores as the context length increases.
\vspace{-1mm} 
\section{Related Works}
\vspace{-1mm}

\subsection{Long Context LLMs}
Enhancing the long-context processing ability of LLMs poses significant challenges for both training and inference due to computational resource constraints.
One line of research aims to scale the position embedding. Based on RoPE~\cite{su2024roformer}, Positional Interpolation~\cite{chen2023extending}, NTK-RoPE~\cite{ntkawarerope}
 are proven to be effective approaches to extend the context length. 
In addition to modifying the positional embedding, ALiBi~\cite{press2021train} and KERPLE~\cite{chi2022kerple} explore the way to encode positional information in attention bias.
Another research direction focuses on devising efficient attention mechanisms to mitigate computational demands. Novel attention mechanisms~\cite{peng2023rwkv,xiao2023efficient, han2023lm, chen2023longlora,ding2023longnet} are proposed to reduce the time complexity and space complexity of standard self-attention.
While Flash Attention~\cite{dao2023flashattention, dao2022flashattention} and Paged Attention~\cite{kwon2023efficient} optimize attention computations by tackling the memory bottleneck while maintaining the precision of attention kernel calculations.

\vspace{-2mm}
\subsection{Evaluation for Long-Context LLMs}
Research work of long context modeling predominantly adopt perplexity as the evaluation metric~\cite{beltagy2020longformer,sun2021long, press2021train, chen2023longlora, peng2023yarn,ntkawarerope}. 
Synthetics tasks, such as retrieval tasks, are used to assess and analyze the ability to model long input for LLMs~\cite{chen2023extending,li2023long}.
However, as discussed in \cite{sun2021long, longllama2}, the perplexity value and performance on synthetic tasks may not adequately reflect a language model's capability in addressing tasks in real-world scenarios.  
Recently an English benchmark~\cite{an2023leval} are proposed for the evaluation of \lllm. \citet{bai2023longbench} introduces a bilingual benchmark, but the quantity of test examples for Chinese is quite limited. Besides the targeted language, \methodname~differs from them in these aspects: (1) It includes novel tasks that closely simulate real-world LLM usage scenarios, and (2) The test samples possess a wider span of context lengths.

%


\vspace{-2mm} 

\section{Conclusion}
\vspace{-1mm}

We presented \methodname, a benchmark for Chinese \lllm, which contains 7 tasks and 7,267 examples. To the best of our knowledge, \methodname~is the first benchmark in this setting. Based on two basic capabilities for \lllm, \ie, information acquisition and reasoning, we collected corresponding tasks and datasets for a comprehensive evaluation.
We benchmarked 8 \lllm~ and provided an in-depth analysis regarding each fine-grained capability.

\section*{Limitations}
\methodname~is specifically crafted for the evaluation of Chinese \lllm. Therefore, it is inapplicable to the LLMs primarily focused on other languages. However, we anticipate that the proposed evaluation framework could provide insights for the construction of benchmarks in other languages. 

In dataset construction, we have tried to gather a broad and varied set of tasks covering all evaluation aspects. Nonetheless, certain tasks, such as code completion and mathematical reasoning, which extend beyond the scope of the Chinese language, are not included. Given these tasks are already supported by a wealth of mature evaluation datasets, we recommend that users employ both them and \methodname~concurrently for model testing.

Furthermore, we adopt matching-based metrics in automatic evaluation, which possess inherent limitations in accurately reflecting the generation quality. We leave the investigation of alternative automatic evaluation methods that have higher alignments with human judgment for future exploration.

\section*{Acknowledgements}
The work described in this paper was partially supported by Laboratory for AI-Powered Financial Technologies, InnoHK initiative and The Government of the HKSAR. The work described in this paper was also partially supported by the Research Grants Council of the Hong Kong Special Administrative Region, China (CUHK 14222922, RGC GRF 2151185).

\bibliography{custom}
\appendix
\section{Supplemential Analysis}\label{app:analysis}
\subsection{Analysis on \textit{StNlab} and \textit{StTDet}}
\label{sec: heat_map}
On \textit{StNLab} and \textit{StTDet}, we use the Average Accuracy (AvgAcc) metric for scoring. On \textit{StNab}, this metric means that we first calculate the proportion of correctly classified news in a context to the total number of news articles in that context, and then take the average across all contexts (\ie, samples). We can also use another alternative metric denoted as Acc, which calculates the total number of correctly classified news articles in each context and sums them up across all contexts, then divides them by the total number of news articles in all contexts. This metric aligns with the Acc metric used on \textit{NLab}. Table \ref{tab:more_results} reports the differences between these two metrics on \textit{StNlab} and \textit{StTDet}. Generally, if an LLM performs well in analyzing short texts but struggles with long texts, the AvgAcc metric will be higher than the Acc metric. The opposite can also occur, although it is less common.

\begin{table}[htbp]
    \centering
    \begin{tabular}{cccc}
    \toprule
    \textbf{Task}    &\textbf{Small}   &\textbf{Medium}   &\textbf{Large}  \\
    \midrule
    LStQA   &100        &100        &100        \\
    LCvMem  &100        &100        &100        \\ 
    LStSum  &400        &400        &800        \\
    StNLab  &800        &800        &1500       \\ 
    StTDet  &400        &800        &800        \\ 
    KpRet   &400        &400        &400        \\ 
    TblQry  &50         &50         &50         \\
    \bottomrule
    \end{tabular}
    \caption{The values of maximum output token limits for the small, medium, and large set in the inference stage.}
    \label{tab:max_tokens}
\end{table} 

Moreover, the format output accuracy (FmtAcc) of \textit{StNLab} and \textit{StTDet} is assessed.
FmtAcc indicates whether the model output, after undergoing our post-processing, conforms to our predefined output. If it does, the value is 1; otherwise, it is 0. On \textit{StNLab}, open-source models have low FmtAcc, leading to lower scores. On \textit{StTDet}, open-source models achieve comparable FmtAcc (\eg, ChatGLM3-6B with 84.0), but still struggle with low AvgAcc, highlighting the challenging nature of typo detection itself.

\begin{table*}[!t]
    \centering
    \begin{tabular}{|l|c|c|c|c|c|c|c|c|}
    \toprule
    \multirow{2}{*}{Model}  &\multicolumn{3}{c|}{\textbf{StNLab}} &\textbf{NLab}   &\multicolumn{3}{c|}{\textbf{StTDet}} &\textbf{TDet} \\ 
                        ~&FmtAcc  &AvgAcc     &Acc    &Acc    &FmtAcc &AvgAcc     &Acc    &Acc    \\ 
    \midrule
     Qwen-7B         &22.11 &4.31   &2.43     &80.91  &0    &0      &0      &18.67 \\ 
     Zh-Alpaca2-7B   &20.13 &4.97   &1.83     &60.09  &6.36    &0.09   &0      &22.27 \\
     ChatGLM3-6B     &0.66 &0.46   &0.24     &86.71  &84.00    &0.91   &0.91      &34.23 \\
     InternLM2-7B    &29.70 &16.46  &4.93     &85.87  &46.72    &2.27   &2.27      &56.20 \\ 
     InternLM2-20B   &18.48 &11.16  &3.43     &84.04  &64.18    &0.91   &0.91      &56.90 \\
     Moonshot-v1-32K  &\textbf{99.10}      &\textbf{89.01}      &\textbf{89.09}     &86.71           &72.36      &25.36       &25.30     &62.06      \\
     GPT-4-Turbo-128K &91.42    &79.70  &73.71    &\textbf{90.31}   &\textbf{92.11}      &\textbf{38.22}   &\textbf{38.01}     &\textbf{75.63}     \\ 
    \bottomrule
    \end{tabular}
    \caption{Peformance (\%) of different metrics in the small set .}
    \label{tab:more_results}
\end{table*}
\begin{figure*}[!t]
    \centering
    \includegraphics[width=1.0\textwidth]{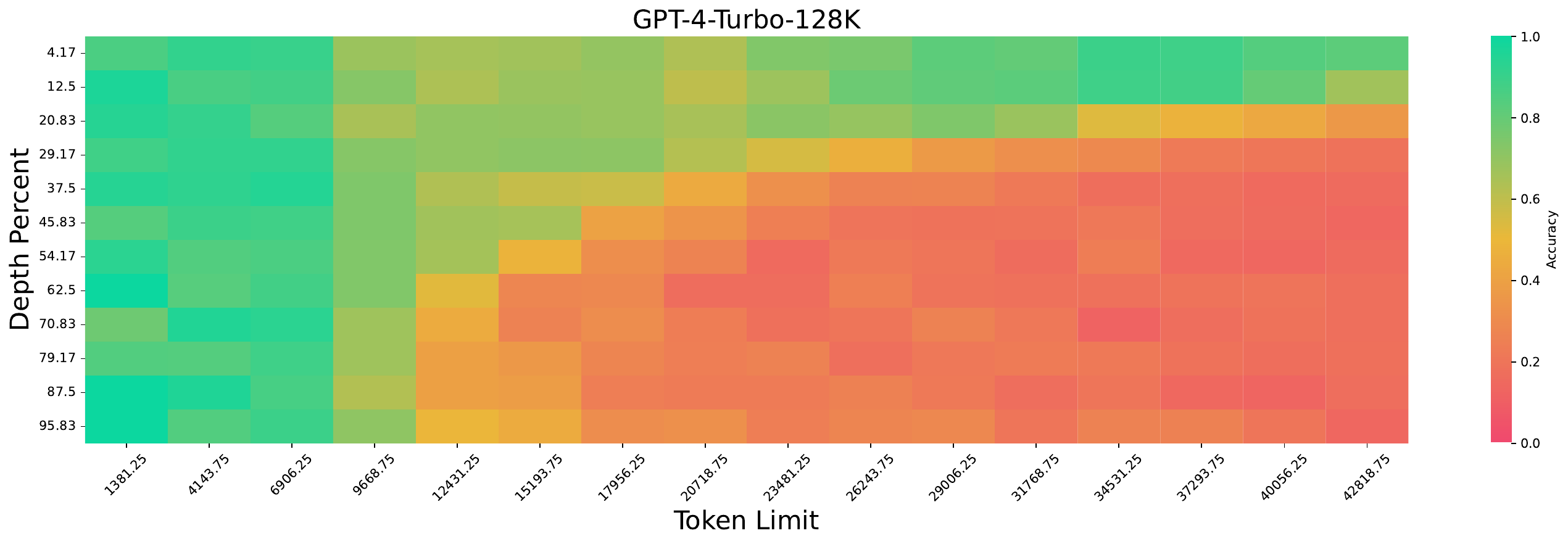}
    \caption{Performance change analysis of GPT-4-Turbo on \textit{StNLab}.}
    \label{fig:GPT-4news}
\end{figure*}
\begin{figure*}[!t]
    \centering
    \includegraphics[width=1.0\textwidth]{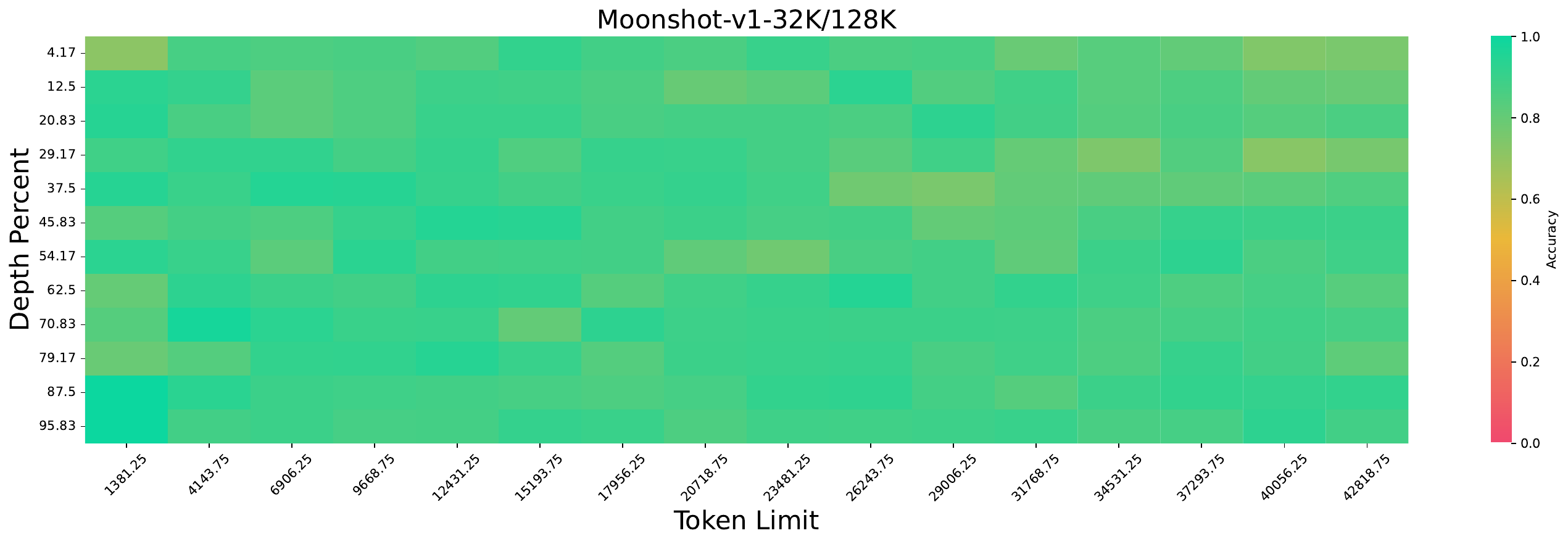}
    \caption{Performance change analysis of Moonshot-v1 on \textit{StNLab}.}
    \vspace{-2mm}
    \label{fig:kimicha-tnews}
\end{figure*}
We also draw heatmaps to analyze the impact of changes in context length and the position depth of news on LLMs' classification accuracy on \textit{StNLab}. The analysis is conducted on all samples from the small set and medium set of \textit{StNLab}. Specifically, for each news within a single context, we first calculate the start and end positions of that news. We then take the average of these two positions as the news depth within the context and discretize it into 12 intervals (\ie, the y-axis). At the same time, we discretize the length of all contexts into 16 intervals (\ie, the x-axis). For each of the 12 intervals on the x-axis and y-axis, we use the midpoint of that interval to represent it. This way, we obtain the discrete position interval and the corresponding discrete context length interval for each news sample. Each news sample is given a classification score of 1.0 if classified correctly, and 0 if classified incorrectly. Afterward, we generate the heatmaps shown in Figure~\ref{fig:GPT-4news} and Figure~\ref{fig:kimicha-tnews} through average aggregation. Figure~\ref{fig:GPT-4news} demonstrates that with an increase in context length, GPT-4-Turbo can only correctly classify news that is closer to the beginning, while making a large number of errors for news located towards the end. Figure~\ref{fig:kimicha-tnews} shows that Moonshot-v1 can classify news with very high accuracy, regardless of their position depth and context length.

\subsection{Analysis on \textit{LCvMem}}
Despite Chinese-Alpaca2-7B generally achieving higher scores than Chinese-LLama2-7B on most datasets, we observe a significant difference on \textit{LCvMem}, where Chinese-Alpaca2-7B's scores are noticeably lower (29.34 vs. 41.10 in the small set). We manually analyze 100 test samples from the small set of \textit{LCvMem} to assess the accuracy of responses from Chinese-Alpaca2-7B and Chinese-LLama2-7B. The analysis reveals that Chinese-Alpaca2-7B achieves an accuracy rate of 61\%, while Chinese-LLama2-7B scored slightly higher at 63\%, indicating no significant distinction. However, given that \textit{LCvMem} is a dataset comprising user dialogues, Chinese-Alpaca2-7B trained by instruction fine-tuning may tend to generate longer responses, leading to a decrease in the F1 score. 


\vspace{-2mm}

\section{Data Samples}
\vspace{-2mm}
\label{sec:data_sample}
For each of the 7 tasks, we show an example of test samples starting on the next page. For Long Story QA, Long Conversation Memory, Key-Passage Retrieval, and Table Querying which require partial-context understanding, the input consists of the prompt, context, and the partial-context-specific question. For Long Story Summarization, Stacked News Labeling, and Stacked Typo Detection which require full-context understanding, the input consists of the prompt and the context.

\clearpage
\clearpage
\begin{tcolorbox}[width=\textwidth, colback=white, colbacktitle=gray!50!white, colframe=gray,title=\textbf{Long Story QA}, coltitle=black]
\label{example:lstqa}
\noindent 
\begin{CJK}{UTF8}{gbsn}
\noindent Prompt: 下面是一部小说的节选。请阅读该小说节选，并尽可能用简洁的短语（或短句）回答给定的问题，不要提供任何解释。小说节选如下：/ Below is an excerpt from a novel. Please read this novel excerpt and answer the given question in as concise a phrase (or sentence) as possible without providing any explanation. Excerpts from the novel are as follows: \\
\\
...... \\ 
乔茂暗道一声：“惭愧!”容得两个转过墙角，相去已远；乔茂连忙窜上房去，向四外一瞥。然后攀垣窜房，走壁爬坡，如飞也似地赶到篱笆门的邻舍房上。不敢探险，且先找着藏身之所，\textcolor{cyan}{然后挨到那两个夜行人现身的所在}，往下面一望：却是一户寻常的乡农之家，一段竹篱，三间北房，两间西房，很宽敞的大院落，院角有一道井栏。试窥看那几间草舍的窗棂，依然是黑沉沉，没有一点灯光，并且也听不见什么声息。这房舍如此的狭窄，又这么悄静，决不象有什么事故发生的样子；乔茂不由诧异起来。 九股烟乔茂久涉江湖，查勘盗踪，足有十二分的把握；只要一入目，便可猜断出十之八九来。看这个草舍，分明不象劫镖强人潜踪之所，更不象梁上君子作案之地，何故竟有两个夜行人窜出呢？乔茂试用一块碎砖，投了一下，也不见动静。当下乔茂提起精神，从邻舍轻轻窜过来，来到院内，仔细查看。先倾耳伏窗，只听得屋内鼾声微作。更验看门窗，的确不象有夜行人出没。\textcolor{cyan}{然后到院内各处一巡，这才来到井栏旁边，发现井旁有只水桶，里面水痕未干，地上也有一片水迹，这分明是刚从井里打完水的情形。
乔茂暗暗点头道：“哦，这就是了。”看这乡农人家，深睡正浓，何来半夜打水？打水的必是刚才那两个夜行人，那么贼人的落脚之处可想而知了}。九股烟乔茂将水桶提了，也向井中打出一些水，喝了一气。随又放下，立刻“嗖”的窜上房来，向村后急打一望。连忙重翻身，窜到街心，施展夜行术，鹿伏鹤行，膝盖碰胸口，脚尖点地面，如星驰也似，投向村后追将过去；那两个夜行人已不知去向。到得村后，正是一带丛林，数畦麦田，通着两条路。乔茂略一端详，择了一条大路，直追下去。\\
...... \\ 
\\
请尽可能简洁地回答下列问题，不要提供任何解释。/ Please answer the following question as succinctly as possible without providing any explanations.\\
问题：小说第七章，乔茂是如何确认夜行人刚才在井旁的？/ Question: In Chapter 7 of the novel, how did Qiao Mao confirm that Night Walker was at the well just now? \\ 
\ \\ 
\textbf{答案}：发现井旁有只水桶，里面水痕未干，地上也有一片水迹。/ \textbf{Translated Ground Truth Reference}：He found a bucket next to the well. The water in it was still wet, and there was also a water stain on the ground.\\ 
\end{CJK}
\end{tcolorbox}
\label{lstqa}

\clearpage
\begin{tcolorbox}[width=\textwidth, colback=white, colbacktitle=gray!50!white, colframe=gray,title=\textbf{Long Conversation Memory}, coltitle=black]
\label{exmaple:lcvmem} 
\begin{CJK}{UTF8}{gbsn}
\noindent Prompt: 你是一个具有聊天陪伴功能的ai伴侣。下面是用户和你的对话记录, 请你阅读对话记录后尽可能简洁地回答问题，不要提供任何解释。对话记录由多天的对话组成：/ You are an AI companion with chat companion function. The following is a conversation record between the user and you. Please read the conversation record and answer the questions as concisely as possible without providing any explanation. Conversation logs consist of multiple days of conversations: \\

.......\\
以下是2023年04月27日的对话记录：\\ 
.......\\
用户: 哦哦，《星际穿越者》啊，我听说过，听起来挺吸引人的！我得找时间去看看。对了，我最近在学摄影，但是总觉得拍出来的照片差点意思，你有什么摄影小技巧可以分享吗？ \\ 
AI: 摄影是一门艺术，也需要不断的实践和学习。你可以尝试从构图和光线入手，比如使用“三分法”来构图，或者利用自然光来营造氛围。还有，多拍多练总是王道！\\ 

用户: 嗯，说的对！我明天就去尝试一下“三分法”。对了，我还喜欢旅游，最近在规划一次小旅行，想去海边走走。你有没有什么好的海滩推荐？\\ 
AI: 海边总是个不错的选择。如果你喜欢清静一些的地方，可以考虑去三亚的亚龙湾，那里的海水清澈，沙滩柔软，风景如画，应该很适合你。\\

用户: 亚龙湾啊，听起来真不错！我会去查查看的。说起旅游，最让我兴奋的就是能在旅途中遇到各种各样的人，听他们的故事。\\ 
AI: 那确实是旅行中最美妙的部分之一。每个人的故事都是独一无二的，能从中学到很多。你最难忘的旅行经历是什么呢？\\

\textcolor{cyan}{用户: 有一次我去了泰国的清迈，那里的文化和风景让我印象深刻。晚上的夜市，色彩斑斓，各种小吃，还有那里人们的热情，真的让人难以忘怀。}\\
AI: 清迈的夜市是闻名遐迩的，那里的生活节奏和氛围跟大城市完全不同。你提到的小吃，有没有什么特别让你想念的呢？\\

用户: 哦，那边的芒果糯米饭简直绝了！每次回忆起那个味道，我就忍不住想再去一次。对了，你能不能给我推荐一些好听的歌？我喜欢边健身边听音乐。\\
AI: 音乐和运动确实是完美的搭档。根据你的热情阳光的性格，我觉得《Can't Stop the Feeling》这首歌很适合你，它的旋律欢快，能让人在运动时充满活力。 \\
......
\ \\

问题：我在4月27日和你聊到曾经旅行去过的城市，这个城市位于哪个国家？/ Question: On April 27, I talked to you about a city I had traveled to. In which country is this city located?\\ 
\ \\ 
答案：泰国。/ \textbf{Translated Ground Truth Reference}：Thailand.\\ 
\end{CJK}
\end{tcolorbox}

\clearpage
\begin{tcolorbox}[width=\textwidth, colback=white, colbacktitle=gray!50!white, colframe=gray,title=\textbf{Long Story Summarization}, coltitle=black]
\label{exmaple:sum} 
\begin{CJK}{UTF8}{gbsn}
\noindent"Prompt: 下面是一个小说节选，请你阅读后写出相应的摘要，不要输出其他任何内容：/ The following is an excerpt from a novel. Please read it and write the corresponding summary. Do not output any other content:\\

\textbf{*Context*} \\ 

答案：在一场关键战役中，安东尼和卡尼迪乌斯分别指挥着各自的军队，而金牛座也带领着他的部队。战斗激烈，但当克利奥帕特拉在战斗中突然逃走，安东尼也跟随她撤退，导致战斗的失败。他们逃往伯罗奔尼撒半岛，斯卡鲁斯决定跟随他们，而卡尼迪乌斯则选择投奔凯撒。在亚历山大的宫殿中，安东尼在绝望中敦促随从逃离并自责逃跑，而克利奥帕特拉则在她的仆人们的鼓励下试图安慰他。安东尼对克利奥帕特拉的逃跑表示谴责，但最终在她的请求下原谅了她。他们决定派遣孩子的校长向凯撒求和，尽管前途未卜，他们还是决定享受一顿盛宴。/ \textbf{Transalted Ground Truth Reference}: During a key battle, Antony and Canidius commanded their respective armies, while Taurus also led his. The battle was fierce, but when Cleopatra suddenly fled during the battle, Antony followed her in retreat, causing the battle to fail. They fled to the Peloponnese, and Scarrus decided to follow them, while Canidius chose to join Caesar. In Alexander's palace, Antony in despair urged his followers to flee and blamed herself for running away, while Cleopatra, encouraged by her servants, tried to comfort him. Antony condemned Cleopatra's escape but eventually forgave her at her request. They decide to send their children's headmaster to sue Caesar for peace, and despite the uncertain future, they decide to enjoy a feast. \\ 
\end{CJK}
\end{tcolorbox}

\clearpage
\begin{tcolorbox}[width=\textwidth, colback=white, colbacktitle=gray!50!white, colframe=gray,title=\textbf{Stacked News Labeling}, coltitle=black]
\label{exmaple:stnlab}  
\begin{CJK}{UTF8}{gbsn}
\noindent"Prompt: 下面是一个新闻列表。每个新闻属于【体育，娱乐，家居，房产，教育，时政，游戏，科技，财经】这九个类别的某一个类别。请按顺序判断并输出每个新闻的所属类别，输出格式为：“新闻1，类别名1 \textcolor{blue}{\textbackslash n} 新闻2，类别名2 \textcolor{blue}{\textbackslash n} ...... \textcolor{blue}{\textbackslash n}新闻N，类别名N”。新闻如下：/ Below is a list of news. Each news belongs to one of the nine categories [sports, entertainment, home, real estate, education, current affairs, games, technology, finance]. Please judge and output the category of each news in order. The output format is: "News1, Category Name 1 \textcolor{blue}{\textbackslash n} News 2, Category Name 2 \textcolor{blue}{\textbackslash n} ...... \textcolor{blue}{\textbackslash n} News N, Category NameN". The list of news is as follows:\\

新闻1:\\
名流·一品 (论坛 相册 户型 样板间 点评 地图搜索) 项目目前在售，户型面积为：46平米一居，80-98平米两居，112-125平米三居，均价6300元/平米，2012年7月入住。项目位于涿州市范阳中路文化广场西300米交通局东侧....\\ 
\ \\ 
新闻2：\\ 
随着国庆长假的临近，装修高峰也随之而来，业主们也开始忙碌起来了，泡装修论坛取经的、逛建材市场实店考察的，只要能用上的途径业主们都乐此不疲，费尽各种心思就是为了给自己给家人一个幸福温馨的家居空间。如何选购建材，特别是卫浴产品，现代的卫浴间更多的已成为人们放松压力、心灵休憩的场所，购买卫浴产品一定要讲究舒适、环保，以及各种人性化设计，从而营造身心放松的舒适空间。在国庆来临之际，各卫浴品牌也都推出各种促销活动...... \\ 
\ \\ 

新闻3：\\ 
金牌，毫无悬念，“后伏明霞时代”，女子双人3米板迎来了最为辉煌的八年，这个辉煌的缔造者就是郭晶晶和吴敏霞。今天，两位顶尖选手的强强联合让水立方的最高领奖台成为最完美的“告别”舞台。\t2000年悉尼奥运会，女子双人3米板项目首次进入奥运会，刚刚复出的伏明霞与郭晶晶的组合由于规定动作质量不高，加上默契不够，输给了俄罗斯的组合帕卡琳娜和伊莲娜......\\
\ \\
\textbf{答案 (Ground Truth Answer)}："新闻1：房产 \textcolor{blue}{\textbackslash n} 新闻2：家居 \textcolor{blue}{\textbackslash n} 新闻3：体育" \\

\end{CJK}
\end{tcolorbox}

\clearpage
\begin{tcolorbox}[width=\textwidth, colback=white, colbacktitle=gray!50!white, colframe=gray,title=\textbf{Stacked Typo Detection}, coltitle=black]
\label{exmaple:sttdet}  
\begin{CJK}{UTF8}{gbsn}
\noindent Prompt: 下面是一部小说的多个段落，每个段落包含段落ID以及段落内容。有些段落中不存在错别字，有些段落有且仅有一个错别字，该错别字与原来的正确汉字同音。请按照段落顺序输出错别字，输出格式为：段落ID，错别字，正确字。不同段落ID用换行符隔开。例如，一个示例输出是"4，蓐，如 \textcolor{blue}{\textbackslash n} 8，颇，坡"。该示例输出中的第一行表示在ID为4的段落中，错别字为蓐，正确字应为如；第二行表示在ID为8的段落中，错别字为颇，正确字应为坡。小说的多个段落如下： / Below are multiple paragraphs of a novel, each paragraph contains the paragraph ID and paragraph content. Some paragraphs have no typos, and some paragraphs have only one typo, which has the same pronunciation as the original correct Chinese character. Please output the typos in the order of the paragraphs. The output format is: paragraph ID, typos, and correct words. Different paragraph IDs are separated by line breaks: \\
\ \\ 
段落ID：1\\
于长水一边发动党团员加紧挖土搬石头，一边帮着石匠钻炮眼崩石崖。土渠开得快，给人们增加了\textcolor{red}{鑫}心；石头崩得响，压倒了庙里的钟鼓。跪香的青壮年在不值班的时候，也溜出庙来参加开渠；老头们说他们心不诚，妨碍了求雨的效果。 \\

段落ID：2\\
两天之后，开渠遇上了新困难：上半截土渠已经挖到庙下边的石崖边，可是石崖上的石头太硬，两天才崩了一排鸡窝窝。原来的估计不正确，光这一段五十尺长五尺深的石渠，一个月也开不过去。这时候退坡的，说闲话的慢慢多起来，也有装病的，也有说家里没吃的不能动的，也有不声不响走开不来的；剩下的人，有的说“一年也开不过去”，有的说：“现在旱得人心慌，还不如等到冬天再开”……原来在庙里跪香的仍回去跪香，原来只在上下工时候去磕个头的也正式编人跪香的班次。\\ 
\ \\ 
段落ID：3 \\
河边人少了，崩开了的石头没人搬，炮声暂且停下来。于长水一边仍叫党团员们搬着石头支持场面不让冷了场，一边脱了鞋，卷起裤管，过到河的对岸，坐在一块石头上，对着这讨厌的石崖想主意。这时候，田里的苗白白地干着，河里的水白白地流着，庙里的钟鼓无用地响着，他觉着实在不是个好滋味。他下了个决心说：“要不能把这么现成的水引到地里去，就算金斗坪没有党！”在火海一样的太阳下，他坐在几乎能烫焦了裤子的石头上，攒着眉头，两眼死盯在这段石崖上，好像想用他的眼光把这段石崖烧化了一样，大约有点把钟没有转眼睛，新办法就被他想出来了。他想要是从石崖离顶五尺高的腰里，凿上一排窟窿，钉上橛子，架上木槽，就可以把水接过去。他这样想着，\textcolor{red}{颢}像已经看见有好几段连在一起的木槽横在这石崖的腰里，水从木槽里平平地流过去，就泻在村北头的平地上。他的眉头展开了。 \\ 
\ \\ 
\textbf{答案 (Ground Truth Answer)}："1, 鑫，信 \textcolor{blue}{{\textbackslash n}} 3，颢，好”。 

\end{CJK}
\end{tcolorbox}

\clearpage
\begin{tcolorbox}[width=\textwidth, colback=white, colbacktitle=gray!50!white, colframe=gray,title=\textbf{Key-Passage Retrieval}, coltitle=black]
\label{exmaple:kpret}  
\begin{CJK}{UTF8}{gbsn}
\noindent Prompt: 请提取下面 JSON 对象中指定键对应的值。只输出对应键的值，不输出任何其他文字。JSON数据如下。 / Please extract the value corresponding to the specified key in the JSON object below. Only the value of the corresponding key is output, and no other text is output. The JSON data is as follows: \\ 
\ \\ 
\{ \\ 
...... \\ 
"9k1PyHdRje6nV4WVddvlGKpCX3Ya9yUA": "《安般守意经》，东汉安世高所译之坐禅佛典，收于《大正藏》经集部，论述以观察呼吸作为修习禅定的方法。此安那般那念与不净观合称二甘露门，为后汉到东晋时期所流行的禅法。康僧会、支愍度、谢敷、支遁、释道安等人都曾为之作注，但已佚失。", \\ 
"9EZ5B8btsIDxeQts8NPLnIMRPhuZ4eys": "答：举世闻名的自由女神像，高高地耸立在纽约港口的自由岛上，象征着美国人民争取自由的崇高理想。自由神像重45万磅，高46米，底座高45米，是当时世界上最高的纪念性建筑，其全称为“自由女神铜像国家纪念碑”，正式名称是“照耀世界的自由女神”。",\\ 
...... \\ 
"\textcolor{cyan}{Smy23rC0V4Dj4NvwbyKXIXeSKmSFgz0n}": "西周时期的各种手工业生产较前有了很大的发展,开始设立职位对各种手工业进行管理.原始瓷器的烧制工艺,在商代后期的基础上有了新的发展和提高,而且出产的地区也较前更为扩大了.在这一阶段,对我国制陶手工业产生重大影响的是已开始把陶器的应用扩大到建筑方面,如板瓦、筒。" \\
...... \\ 
\} \\ 
\ \\ 
键 (Key)：Smy23rC0V4Dj4NvwbyKXIXeSKmSFgz0n \\
值 (Value)：\\
\ \\ 
\textbf{答案 (Ground Truth Answer)}: 西周时期的各种手工业生产较前有了很大的发展,开始设立职位对各种手工业进行管理.原始瓷器的烧制工艺,在商代后期的基础上有了新的发展和提高,而且出产的地区也较前更为扩大了.在这一阶段,对我国制陶手工业产生重大影响的是已开始把陶器的应用扩大到建筑方面,如板瓦、筒。
\end{CJK}
\end{tcolorbox}

\clearpage
\begin{tcolorbox}[width=\textwidth, colback=white, colbacktitle=gray!50!white, colframe=gray,title=\textbf{Table Querying}, coltitle=black]
\label{exmaple:tblqry}  
\begin{CJK}{UTF8}{gbsn}
\noindent Prompt: 请根据下列多个表格的内容回答给定问题，表格为Markdown格式。请直接返回问题的答案，除此之外不要输出其他内容。表格如下： / Please answer the given questions based on the contents of the following multiple tables. The tables are in Markdown format. Please return the answer to the question directly and do not output anything else. The tables are as follows: \\ 
\ \\ 
...... \\ 
表格-191924：\\ 
主场|成立|教练|团队|地点 \\ 
Athelstone Recreation Reserve|1989|Unknown|国民银行|Athelstone \\ 
Byrne Park|1951|Anthony Brevi|北方恶魔|Port Pirie \\ 
O'Sullivan Beach Sports Complex|1997|Aldo Maricic|南阿德莱德|O'Sullivan Beach \\ 
Karingal Reserve|1970|Ben Dale|西福德|Seaford \\ 
Karbeethan Reserve|1978|John Duthie|高勒|Evanston \\ 
A A Bailey Recreation Ground|2011|Alan Paice|斯图尔特雄狮足球俱乐部|Clarence Gardens \\ 
...... \\ 

表格-171426：\\
结果|首次当选|候选人|区|派对|现任 \\ 
Re-elected|1982|Ronald D. Coleman (D) 57.4\% Jack Hammond (R) 42.6\%|德克萨斯州 16|Democratic|Ronald D. Coleman \\
Lost renomination Democratic hold|1966|Albert Bustamante (D) Unopposed|德克萨斯州 23|Democratic|Abraham Kazen, Jr. \\
Re-elected|1976|Sam B. Hall (D) Unopposed|德克萨斯州 1|Democratic|Sam B. Hall \\ 
Lost re-election Republican gain|1982|Dick Armey (R) 51.3\% Tom Vandergriff (D) 48.7\%|德克萨斯州 26|Democratic|Tom Vandergriff \\ 
Re-elected|1961|Henry B. Gonzalez (D) Unopposed|德克萨斯州 20|Democratic|\textcolor{cyan}{Henry B. Gonzalez} \\ 
Re-elected|1978|Martin Frost (D) 59.5\% Bob Burk (R) 40.5\%|德克萨斯州 24|Democratic|Martin Frost \\ 
Re-elected|1980|Jack Fields (R) 64.6\% Don Buford (D) 35.4\%|德克萨斯州 8|Republican|Jack Fields \\ 
.....  \\ 
\ \\ 
问题：在表格-171426中，当“区”这一列的值为“德克萨斯州 20”时，“现任”这一列的值是多少？/ Questionos:  In table-171426, when the value of the "区" column is "德克萨斯州 20", what is the value of the "现任" column? \\ 

\ \\ 
\textbf{答案 (Ground Truth Answer)}: Henry B. Gonzalez  \\

\noindent 
\end{CJK}
\end{tcolorbox}

\end{document}